# Comparative Study and Optimization of Feature-Extraction Techniques for Content based Image Retrieval

Aman Chadha
Department of Electrical and Computer Engineering
University of Wisconsin-Madison

Sushmit Mallik
Department of Electrical and Computer Engineering
North Carolina State University

Ravdeep Johar
Department of Computer Sciences
Rochester Institute of Technology

## ABSTRACT

The aim of a Content-Based Image Retrieval (CBIR) system, also known as Query by Image Content (QBIC), is to help users to retrieve relevant images based on their contents. CBIR technologies provide a method to find images in large databases by using unique descriptors from a trained image. The image descriptors include texture, color, intensity and shape of the object inside an image. Several feature-extraction techniques viz., Average RGB, Color Moments, Co-occurrence, Local Color Histogram, Global Color Histogram and Geometric Moment have been critically compared in this paper. However, individually these techniques result in poor performance. So, combinations of these techniques have also been evaluated and results for the most efficient combination of techniques have been presented and optimized for each class of image query. We also propose an improvement in image retrieval performance by introducing the idea of Query modification through image cropping. It enables the user to identify a region of interest and modify the initial query to refine and personalize the image retrieval results.

## 1. INTRODUCTION

With the recent outburst of multimedia-enabled systems, the need for multimedia retrieval has increased by leaps and bounds. Due to the complexity of multimedia contents, image understanding is a difficult-albeit-interesting topic of research, within the domain of multimedia retrieval. Extracting valuable knowledge from a large-scale multimedia repository, usually referred to as “multimedia mining”, has recently caught up as a domain of interest amongst researchers.Typically, in the development of an image requisition system, semantic image retrieval relies heavily on the related captions, e.g., file-names, categories, annotated key-words, and other manual descriptions.Searching of images is predominantly based upon associated metadata such as keywords, text, etc.The term CBIR describes the process of retrieving desired images from the large collection of database on the basis of features that can be automatically extracted from the images.The ultimate goal of a CBIR system is to avoid the use of textual descriptions in the hunt for an image by the user. Unfortunately, this kind of a textual-based image retrieval system always suffers from two problems: high-priced manual annotation and inaccurate and inconsistent automated annotation. On one hand, the cost associated with manual annotation is prohibitive with regards to a large-scale data set. On the other hand, inappropriate automated annotation yields distorted results for semantic image retrieval. As a result, a number of powerful image retrieval algorithms have been proposed to deal with such problems over the past few years. CBIR is the mainstay of current image retrieval systems.

In CBIR, retrieval of image is based on similarities in their contents, i.e., textures, colors, shapes etc., which are considered the lower level features of an image. These conventional approaches for image retrieval are based on the computation of the similarity between the users query and images. In CBIR each image stored in the database, has its features extracted and compared to the features of the query image. Thus, broadly, it involves two processes, viz,feature extraction and feature matching [8].

Feature extraction involves the image features to a distinguishable extent. Average RGB, Color Moments, Co-occurence, Local Color Histogram, Global Color Histogram and Geometric Momentsare used to extract features from the test image. Feature matching, on the other hand, involves matching the extracted features to yield results that exhibitvisual similarities.

Feature vectors are calculated for the given image. The Euclidean distance is used as default implementation for comparing two feature vectors. If the distance between feature vectors of the query image and images in the database is small enough, the corresponding image in the database is to be considered as a match to the query. The search is usually based on similarity rather than on exact match and the retrieval results are then ranked accordingly to a similarity index. Figure 1 shows the block diagram of a basic CBIR system.

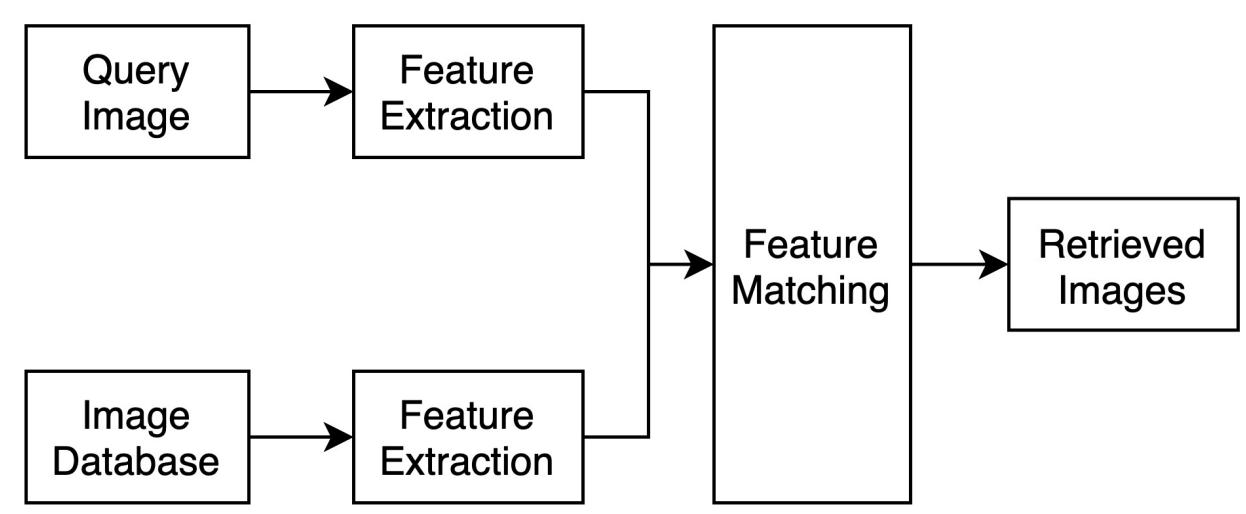

**Fig1: Block diagram of a basic CBIR system**

This paper discusses the detailed comparison of different feature extraction techniques on several groups (classes) of images. This remainder of this paper is organized as follows. Section 2 briefly describes the review of literature. Feature extraction techniques have been discussed in section 3. Section 4 describes the standard database used for the comparative analysis. Section 5 and section 6 describe the general methodology and the parameters used for comparison. A comparative analysis of the performance of each technique on various classes of images has been put forth in section 7. Optimization for each class of images is carried out and the most efficient combination of techniques is presented in Section 8. Section 9 puts forth the feature of user query modification via cropping and discusses the results and applications in section. Finally, conclusion of the paper and presenting future research directions are in section 10.

# 2. REVIEW OF LITERATURE

Querying by Image Content (QBIC) was developed by IBM to retrieve images without any verbal description, but by sorting the image database and querying it by shape, color, texture and spatial location. The application of image processing and related techniques to derive retrieval features is referred to as Content-Based Image Retrieval (CBIR). Web Seek, PicToSeek, NECAMORE, UCSB NeTra and Image Rover are web media search engines that follow the 'query by similar image' paradigm. Virage Video Engine [2] was developed for multimodal indexing and retrieval of videos. Library-based coding [3] is a way of representing images and uses retrieval-enabled MPEG for efficient querying and retrieval.

ElasticElastic matching of images [1] for sketch-based IR, windowed search over location and scale for objects based IR and fractal block code based image histograms for texture based IR systems were proved as the efficient retrieval system. Color correllogram takes into consideration the spatial distribution of colors and thus is an enhancement over previous methods. A stochastic model like Photo book and blob world system, analyses images in both time and frequency domain using 2D discrete wavelet transform and does regular fragmentation of images into homogeneous regions. Daubechies' wavelet transform was then introduced in the Wavelet Based Image Indexing and Searching (WBIIS) system initially to improve the color layout feature. ImageScape is one of the most efficient search engines for finding visual media, uses vector quantization for compressing image databases, and K-D trees for fast search over high-dimensional space.

Semantics-sensitive integrated matching is a wavelet-based approach like the WBIIS system [4], but uses better strategies to capture image semantics, better integrated region matching (IRM) metrics and image segmentation algorithms. FACERET [5] is an interactive face retrieval system which uses self-organizing maps and relevance feedback to solve the complexity with non-trivial high level human description. It uses Principal Component Analysis (PCA) projections to project face images to a dimensionally reduced space. Another approach is the linguistic indexing of pictures [6] using a 2-D multi-resolution hidden Markov model (2DMHMM) for the statistical modeling process and statistical linguistic indexing. Text description is fitted to each image and that will describe the relationship between clusters of image and clusters of feature vectors at multi resolution

Field Programmable Gate Arrays (FPGA) enabled efficient ways of retrieving images with a network of imaging devices. CORNITA enabled image retrieval on World Wide Web (WWW) using query based on keyword, images and relevance feedback. Ontological Query Language (OQUEL) was introduced introduced for querying of images using ontology which provides a language framework with grammar and extensible vocabulary. Personalizable Image Browsing Engine (PIBE) uses browsing tree, a hierarchical browsing structure for quick search and visualization of large image collections and Costume (2005) enabled automatic video indexing.

Evolutionary searching (2000), feature dependency measure (2002), boosting (2004) and Bayes' error (2005) were proposed for generic feature selection. Support Vector Machine (SVM), a swiftly growing field within pattern recognition-based feature detection, is used for facial recognition system. MultiMediaInformation Retrieval (MMIR) enabled image retrieval using Informix data blades, IBM DB2 extenders and Oracle cartridges. An IR framework called OLIVE (2008) provides dual access to web images and used Google images and PIRIA visual search engines. A new graph-based link analysis technique called Imagination makes use of accurate image annotation.

Over the years, several efficient algorithms in CBIR shed light on new interesting facts on multimedia, computer vision, information retrieval and human-computer interaction. It has resultant in a high resolution, high-dimension and maximum throughput of images searchable by the content. Due to its high resolution and quality of the image retrieved, its application is expanded in the field of biomedical imaging, astronomy and various other scientific fields.

# 3. FEATURE EXTRACTION

## 3.1 Gray Level Co-occurrence Matrix

Gray Level Co-occurrence Matrices (GLCM)is a popular representation for the texture in images. Theycontain a count of the number of times a given feature (e.g., a given gray level)occurs in a particular spatial relation to another given feature.GLCM, one of the most known texture analysis methods, estimate image properties related to second-order statistics. We used GLCM techniques for texture description in experiments with 14 statistical features extracted from them. The process involved is follows:

1. Compute co-occurrence matrices for the images in the database and also the query image.

Four matrices will be generated for each image [6].

2. Build up a 4×4 features form the previous co-occurrence matrices as shown in Table 1

**Table 1.Four main features used in feature extraction**

| Feature | Formula |
| --- | --- |
| **Energy** | $\sum_i \sum_j P^2(i,j)$ |
| **Entropy** | $\sum_i \sum_j P(i,j) \log P(i,j)$ |
| **Contrast** | $\sum_i \sum_j (i-j)^2 P(i,j)$ |
| **Homogeneity** | $\sum_i \sum_j \frac{P(i,j)}{1+\lvert i-j \rvert}$ |

## 3.2 Color Histogram

Color is the most widely used "feature"owing to its intuitiveness compared with other features and most importantly, it is easy to extract from the image. The color histogram depicts color distribution using a set of bins. However, a CBIR system based on color featuresis often found to yield distorted results, because it uses global color feature which cannot capture color distributions or textures within the image in some cases. To improve the preferment of the color extraction we divide the color histogram feature into global and local color extraction.

Using Global Color Histogram (GCH), an imagewill be encoded with its color histogram, and the distance between two imageswill be determined by the distance between their color histograms.

Local color histogram (LCH) can give some sort of spatial information, however the con associated with it is that it uses very large feature vectors.LCH includes information concerning the colordistribution of regions. The first step is to segment the image into blocks and thento obtain a color histogram for each block. An image will then be represented bythese histograms. When comparing two images, we calculate the distance, usingtheir histograms, between a region in one image and a region in same location inthe other image. The distance between the two images will be determined by thesum of all these distances.

However, it does not include information concerning the color distribution of the regions, so the distance between images sometimes cannot show the real difference between images. Moreover, in the case of a GCH, it is possible for two different images to have a very short distance between their color histograms. This is their main disadvantage.

## 3.3 Geometric Moments

In image processing, computer vision and related fields, an image moment is acertain particular weighted average (moment) of the image pixels' intensities, ora function of such moments, usually chosen to have some attractive property orinterpretation. Image moments are useful to describe objects after segmentation.

Simple properties of the image which are found via image moments include area(or total intensity), its centroid, and information about its orientation.This feature use only one value for the feature vector, however, the performance of current implementation isn't well scaled, [7] which means that when the image size becomes relatively large, computation of the feature vector takes a large amount of time. The pros of using this feature combine with other features such co-occurrence, which can provide a better result to user.

## 3.4 Average RGB

The objective of use this feature is to filter out images with larger distance at first stage when multiple feature queries are involved. Another reason of choosing this feature is the fact that it uses a small number of data to represent the feature vector and it also uses less computation as compared to others. However, the accuracies of query result could be significantly impact if this feature is not combined with other features.

## 3.5 Color Moments

To overcome the quantization effects of the color histogram, we use the colormoments as feature vectors for image retrieval. Since any color distribution canbe characterized by its moments and most information is concentrated on the low-ordermoments, only the first moment (mean), the second moment (variance) andthe third moment (skewness) are taken as the feature vectors. With a very reasonable size of feature vector, the computation is not expensive [9].Color Moments are measures that can be differentiate images based on their feature of color, however, the basic concept behind color moments lays in the assumption that the distribution of color in an image can be interpreted as a probability distribution. The advantage is that, its skew-ness can be used as a measure of the degree of asymmetry in the distribution.

# 4. DATABASE USED

We have used a standard database for testing, the WANG database [10], [11]. It is a subset of 1,000 images of the Corel stock photo database, manually selected and which form 10 classes of 100 images each. The 10 classes are African people and villages, beaches, buildings, buses, dinosaurs, elephants, flowers, horses, mountains and glaciers, and food. It can ve visualized as a similar way to a photo retrieval task with several images from each category and a user having an image from a particular category and looking for similar images. The 10 classes are used for relevance estimation. Given a query image, it is assumed that the user is searching for images from the same class, and the remaining 99 images from the same class are considered relevant and the images from all other classes are considered irrelevant. For example, let us assume, the user gives an image from Class 2 as a query, all the images belonging to that class will be considered as relevant and the rest irrelevant. So if there are 60 images displayed in result and 20 of them belong to Class 2, we have 20 relevant and 40 non relevant images.

# 5. METHODOLOGY

We will analyze six techniques one by one using a query image from each class of the WANG database. The six techniques are Average RGB, Color Moments, Cooccurence, Local Color Histogram, Global Color Histogram and Geometric Moment. These six techniques will be evaluated using the parameters, Time, Accuracy and Redundancy Factor, we will be explained in the next section.

The goal is to find the optimum combination of techniques to be used for each class of query which results is the best possible Time, Accuracy and Redundancy Factor, as compared to using any single technique at one point of time. This will result in an 'adaptive' CBIR system, which can adapt itself according to query image given by the user and use the relevant techniques for the image retrieval process to produce the best results.

# 6. PARAMETERS

The parameters, Time, Accuracy and Redundancy Factor, are explained as follows:

## 6.1 Time

It is the time taken in seconds for the retrieval task to complete, at the end of which the system returns the images which are matched with the features of the query images, according to the technique used.

## 6.2 Accuracy

Accuracy of an image retrieval task is defined as the ratio of the number of relevant images retrieved to the total number of images retrieved expressed in percentage.

$$\text{Accuracy} = \frac{\text{Number of relevant images}}{\text{Total number of images retrieved}} \times 100 \qquad (1)$$

Where, total number of images retrieved = number of relevant images + number of irrelevant images

For example, if an image query result in 100 images with 75 relevant images, then the accuracy of the retrieval process is given by:

$$\text{Accuracy} = \frac{75}{100} \times 100 = 75\%$$

We make an assumption to calculate the accuracy values using the first 50 relevant image results for uniformity and simplicity of calculations.Accuracy is a vital parameter for evaluation as it is a direct measurement of the quality and user satisfaction of the image retrieval process.

## 6.3 Redundancy Factor

Redundancy Factor (RF) is one aspect which has been largely neglected in the analysis of CBIR techniques. It is a measure to take into account the extent of irrelevant images returned upon completion of a retrieval process.It is expressed as:

$$RF= \frac{\text{(Total number of images retrieved)} - \text{(Total number of images in a class)}}{\text{Total number of images in a class}} \quad (2)$$

In the Wang database, there are 10 classes of images with 100 images of each class. So, for the Wang database, RF is calculated as:

$$RF= \frac{\text{Total number of images retrieved} - 100}{100} \quad (3)$$

For instance, if the retrieval process returns 125 images, the RF will be calculated as:

$$RF = \frac{(125\text{-}100)}{100} = 0.25$$

Since the Wang database has 1000 images, the RF can vary between -1 to 9. The Ideal RF is obviously 0, which means that all the retrieved images, belong to the same class of the query image. So, if the RF is greater than 0, it means that the system is being over-worked, resulting in excess results. If the RF is less than 0, it means that the system is being under-worked and hence, underperforming. This means, even, if the accuracy might be high, it will reduce the full potential of the retrieval process by eliminating some images from the same class, which could be of some use to the user.

## 7. ANALYSIS

The analysis of the six techniques will be evaluated using the parameters described before. We assume that an accuracy of 50% or more is termed as 'good' performance and less than 50% is termed 'bad' performance.

## 7.1 Average RGB

**Table 2. Results for the Average RGB**

| Image Class | Images retrieved | Time (sec) | Relevant Images | Accuracy (%) | RF |
|---|---|---|---|---|---|
| Class 1 | 151 | 7 | 19 | 38 | 0.51 |
| Class 2 | 6 | 7 | 3 | 50 | -0.94 |
| Class 3 | 136 | 6 | 6 | 12 | 0.36 |
| Class 4 | 213 | 10 | 28 | 56 | 1.13 |
| Class 5 | 170 | 9 | 50 | 100 | 0.70 |
| Class 6 | 53 | 8 | 12 | 24 | -0.47 |
| Class 7 | 20 | 8 | 20 | 100 | -0.8 |
| Class 8 | 85 | 9 | 41 | 82 | -0.15 |
| Class 9 | 22 | 9 | 12 | 54.54 | -0.78 |
| Class 10 | 39 | 7 | 17 | 43.58 | -0.61 |

We can see that there are a lot of negative RF values, which gives an indication that Average RGB is resulting in the system to underperform. Mean Time taken is 8 seconds. Mean Accuracy obtained is 56.01%. Mean RF results in -0.105.

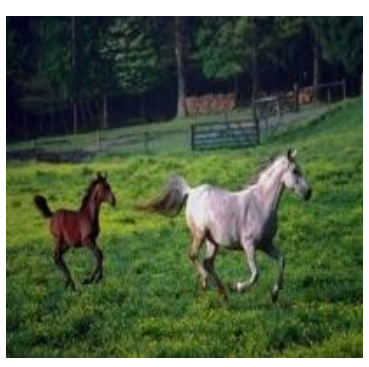
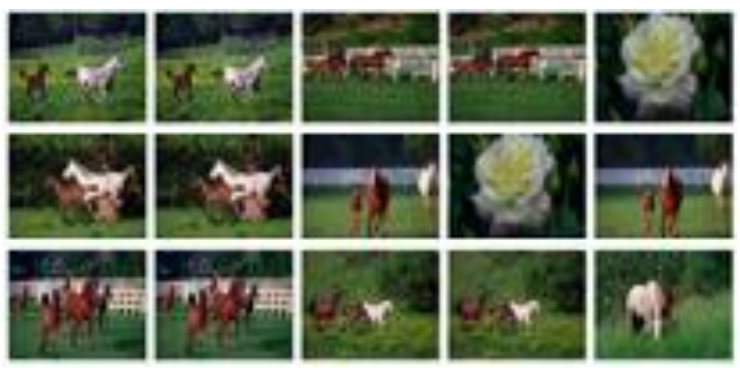

**Fig 2: Illustration of query and retrieval using Average RGB method**

## 7.2 Color Moments

**Table 3. Results for the Color Moments technique**

| Image Class | Images retrieved | Time (sec) | Relevant Images | Accuracy (%) | RF |
|---|---|---|---|---|---|
| Class 1 | 633 | 15 | 12 | 24 | 5.33 |
| Class 2 | 209 | 10 | 22 | 44 | 1.09 |
| Class 3 | 757 | 15 | 5 | 10 | 6.57 |
| Class 4 | 777 | 16 | 16 | 32 | 6.77 |
| Class 5 | 226 | 11 | 50 | 100 | 1.26 |
| Class 6 | 300 | 11 | 12 | 24 | 2.00 |
| Class 7 | 256 | 10 | 42 | 84 | 1.56 |
| Class 8 | 688 | 14 | 42 | 84 | 5.88 |
| Class 9 | 202 | 9 | 20 | 40 | 1.02 |
| Class 10 | 403 | 13 | 21 | 42 | 3.03 |

We can see that Color Moments results in high RF. Mean Time taken is 12.4 seconds. Mean Accuracy obtained is 48.4%. Mean RF results in 3.45.

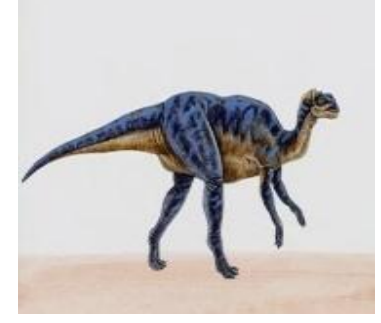
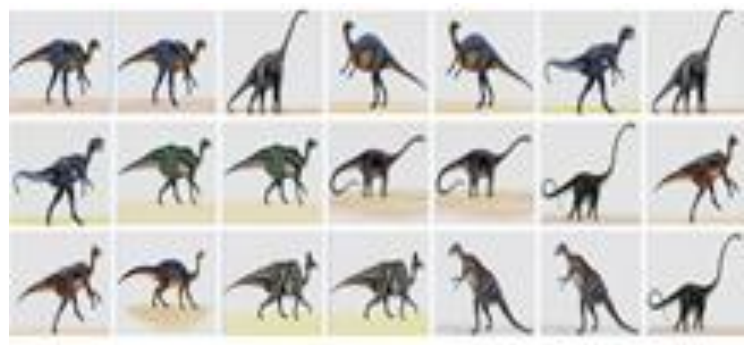

**Fig 3: Illustration of query and retrieval using Color Moments method**

## 7.3 Co-occurence

**Table 4. Results for the Co-occurence technique**

| Image Class | Images retrieved | Time (sec) | Relevant Images | Accuracy (%) | RF |
|---|---|---|---|---|---|
| Class 1 | 25 | 8 | 11 | 44 | -0.75 |
| Class 2 | 51 | 8 | 10 | 20 | -0.49 |
| Class 3 | 44 | 9 | 3 | 6.8 | -0.56 |
| Class 4 | 16 | 9 | 11 | 68.75 | -0.84 |
| Class 5 | 58 | 11 | 50 | 100 | -0.42 |
| Class 6 | 78 | 12 | 7 | 14 | -0.22 |
| Class 7 | 64 | 10 | 45 | 90 | -0.36 |
| Class 8 | 87 | 11 | 33 | 66 | -0.13 |
| Class 9 | 37 | 10 | 9 | 18 | -0.63 |
| Class 10 | 20 | 9 | 3 | 15 | -0.80 |

Mean Time taken is 9.7 seconds. Mean Accuracy obtained is 44.25%. Mean RF results in -0.52.

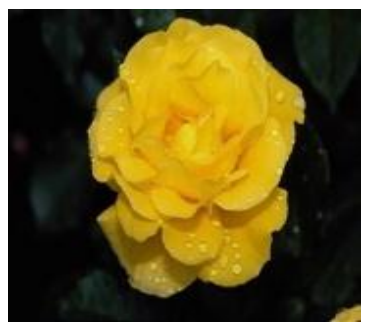

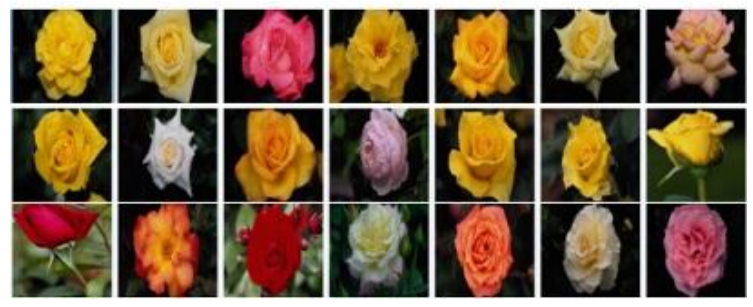

**Fig 4: Illustration of query and retrieval using Co-occurence method**

## 7.4 Local Color Histogram

**Table 4. Results for the Local Color Histogram**

| Image Class | Images retrieved | Time (sec) | Relevant Images | Accuracy (%) | RF |
|---|---|---|---|---|---|
| Class 1 | 327 | 17 | 9 | 18 | 2.27 |
| Class 2 | 50 | 9 | 7 | 14 | -0.50 |
| Class 3 | 18 | 8 | 2 | 11.11 | -0.82 |
| Class 4 | 766 | 21 | 10 | 20 | 6.66 |
| Class 5 | 271 | 13 | 48 | 96 | 1.71 |
| Class 6 | 723 | 15 | 10 | 20 | 6.23 |
| Class 7 | 407 | 13 | 45 | 90 | 3.07 |
| Class 8 | 418 | 11 | 26 | 52 | 3.18 |
| Class 9 | 35 | 8 | 16 | 45.71 | -0.65 |
| Class 10 | 666 | 14 | 14 | 28 | 5.66 |

Mean Time taken is 12.9 seconds. Mean Accuracy obtained is 39.48%. Mean RF results in 2.68.

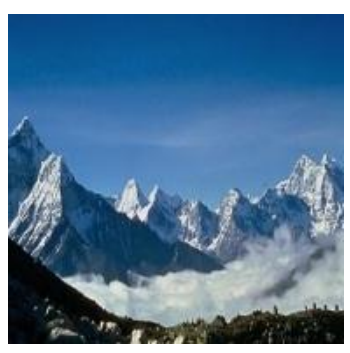

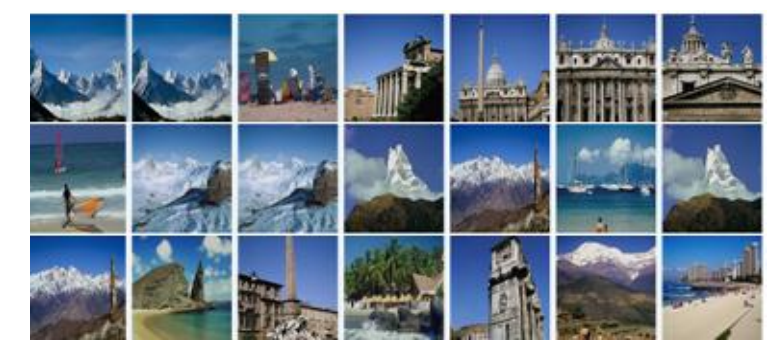

**Fig5: Illustration of query and retrieval using Local Color Histogram method**

## 7.5 Global Color Histogram

**Table 5. Results for Global Color Histogram technique**

| Image Class | Images retrieved | Time (sec) | Relevant Images | Accuracy (%) | RF |
|---|---|---|---|---|---|
| Class 1 | 908 | 16 | 24 | 48 | 8.08 |
| Class 2 | 59 | 9 | 6 | 12 | -0.41 |
| Class 3 | 595 | 14 | 9 | 18 | 4.95 |
| Class 4 | 836 | 16 | 22 | 44 | 7.36 |
| Class 5 | 210 | 10 | 50 | 100 | 1.10 |
| Class 6 | 260 | 11 | 19 | 38 | 1.60 |
| Class 7 | 185 | 10 | 43 | 86 | 0.85 |
| Class 8 | 686 | 14 | 43 | 86 | 5.86 |
| Class 9 | 114 | 9 | 17 | 34 | 1.14 |
| Class 10 | 782 | 15 | 30 | 60 | 6.82 |

Mean Time taken is 12.4 seconds. Mean Accuracy obtained is 52.6%. Mean RF results in 3.73.

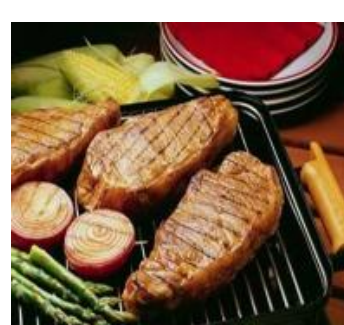

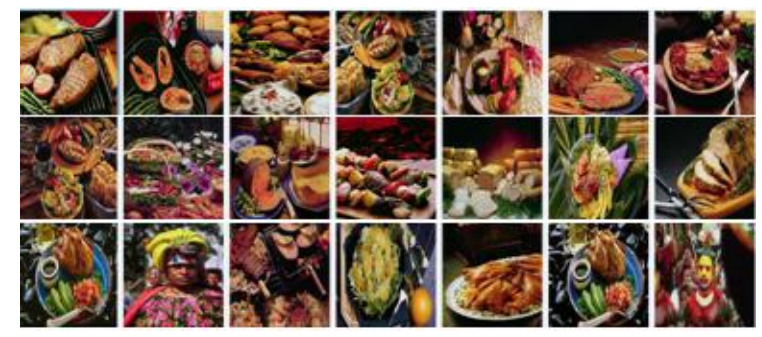

**Fig6: Illustration of query and retrieval using Global Color Histogram method**

## 7.6 Geometric Moment

**Table 6. Results for the Geometric Moment technique**

| Image Class | Images retrieved | Time (sec) | Relevant Images | Accuracy (%) | RF |
|---|---|---|---|---|---|
| Class 1 | 357 | 10 | 5 | 10 | 2.57 |
| Class 2 | 1000 | 21 | 2 | 4 | 9.00 |
| Class 3 | 98 | 10 | 2 | 4 | -0.02 |
| Class 4 | 1000 | 21 | 7 | 14 | 9.00 |
| Class 5 | 447 | 13 | 7 | 14 | 3.47 |
| Class 6 | 1000 | 20 | 5 | 10 | 9.00 |
| Class 7 | 1000 | 19 | 8 | 16 | 9.00 |
| Class 8 | 884 | 17 | 8 | 16 | 7.84 |
| Class 9 | 1000 | 19 | 7 | 14 | 9.00 |
| Class 10 | 915 | 17 | 7 | 14 | 8.15 |

This is the most ineffective technique as it performs poorly in all parameters of evaluation. Mean Time taken is 16.7 seconds. Mean Accuracy obtained is 11.6%. Mean RFis 6.70.

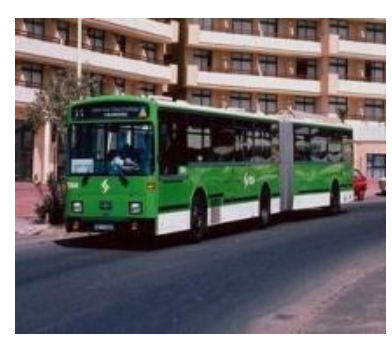

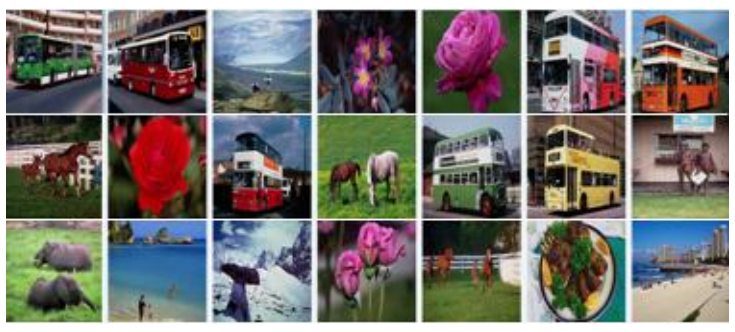

**Fig 7: Illustration of query and retrieval using Geometric Moment method**

## 7.7 Combined Approach

If we calculate the mean time taken, accuracy and RF of all 6 techniques, we find that the mean time is 12.01 seconds, mean accuracy is 42.05% and mean RF is 2.65. Since the individual techniques result in an accuracy of below 50% in most cases, we combine all of them to see the results.

**Table 7. Results for the Combined Approach**

| Image Class | Images retrieved | Time (sec) | Relevant Images | Accuracy (%) | RF |
|---|---|---|---|---|---|
| Class 1 | 2 | 52 | 2 | 100 | -0.98 |
| Class 2 | 1 | 52 | 1 | 100 | -0.99 |
| Class 3 | 1 | 52 | 1 | 100 | -0.99 |
| Class 4 | 6 | 52 | 6 | 100 | -0.94 |
| Class 5 | 10 | 53 | 10 | 100 | -0.90 |
| Class 6 | 7 | 51 | 5 | 71.42 | -0.93 |
| Class 7 | 9 | 50 | 9 | 100 | -0.91 |
| Class 8 | 16 | 50 | 15 | 93.75 | -0.84 |
| Class 9 | 2 | 51 | 2 | 100 | -0.98 |
| Class 10 | 2 | 50 | 1 | 50 | -0.98 |

Mean Time taken is 51.3 seconds. Mean Accuracy obtained is 91.51%. Mean RF results in -0.90. As we can see, the accuracy rates are near perfect for most cases. The drawbacks are poor redundancy and considerably long retrieval time. Hence, we show that combining all techniques results in a substantial rise in the accuracy rates.

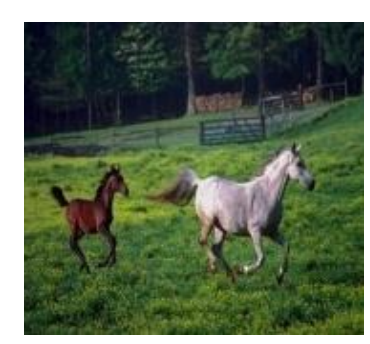

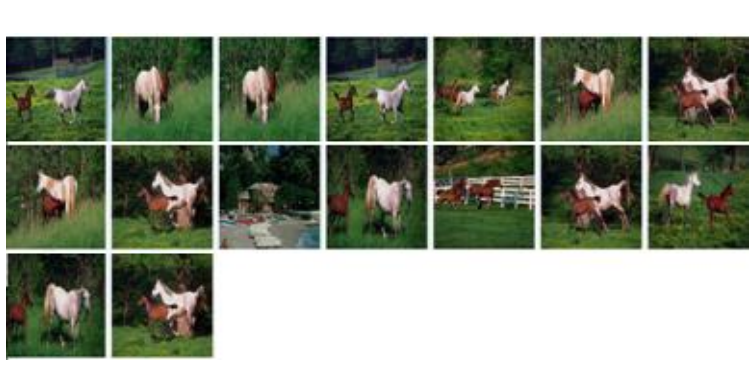

**Fig 8: Illustration of query and retrieval using Combined Approach method**

# 8. OPTIMIZATION

Previously, we saw that the combined approach gives excellent accuracy but with poor RF and long retrieval times. Hence, we optimize the image retrieval process by selecting the most efficient combinations to give best accuracy possible.

## 8.1 Class 1

The best results for Class 1 images are given by using Cooccurence and Global Color Histogram resulting in an accuracy of 64% in 19 seconds with RF of 1.98.

## 8.2 Class 2

The best results for Class 2 images are given by using Average RGB and Color Moments resulting in an accuracy of 100% in 16 seconds with RF of -0.94.

## 8.3 Class 3

The best results for Class 3 images are given by using Local Color Histogram and Global Color Histogram resulting in an accuracy of 57.14% in 17 seconds with RF of – 0.93.

## 8.4 Class 4

The best results for Class 4 images are given by Average RGB and Cooccurance resulting in an accuracy of 100% in 17 seconds with RF of -0.94. Moreover, Co-occurence, Global Color Histogram and Geometric Moment gives 73.33% accuracy in 25 seconds with RF of -0.85 and Average RGB and Geometric Moment gives an accuracy of 54% in 16 seconds with RF of 0.52.

## 8.5 Class 5

The best results for Class 5 images are given by Average RGB and Co-occurence method resulting in an accuracy of 100% in 16 seconds with RF of -0.44.

## 8.6 Class 6

The best results for Class 6 images are given by using Average RGB,Co-occurrence and Global Color Histogram resulting in an accuracy of 77.77% in 25 seconds with RF of -0.91.

## 8.7 Class 7

The best results for Class 7 images are given by using Co-occurrence, Local Color Histogram and Global Color Histogram resulting in an accuracy of 100% in 20 seconds with RF of -0.55.

## 8.8 Class 8

The best results for Class 8 images are given by using Average RGB and Color Moments resulting in an accuracy of 96% in 17 seconds with RF of - 0.16.

## 8.9 Class 9

The best results for Class 9 images are given by using Average RGB, Local Color Histogram and Global Color Histogram resulting in an accuracy of 60% in 26 seconds with RF of -0.90.

## 8.10 Class 10

The best results for Class 10 images is given by using Global Color Moment and Color Moments resulting in an accuracy of 62% in 22 seconds with RF of 2.73.

## 8.11 Improvements

Compiling all the results from all the classes, we can clearly see an improvement in Accuracy, Time taken for retrieval and Redundancy factor.

**Table 8. Results obtained by Optimization**

| Image Class | Images retrieved | Time (sec) | Relevant Images | Accuracy (%) | RF |
|---|---|---|---|---|---|
| Class 1 | 298 | 19 | 32 | 64 | 1.8 |
| Class 2 | 6 | 16 | 6 | 100 | -0.94 |
| Class 3 | 7 | 17 | 4 | 57.14 | -0.93 |
| Class 4 | 6 | 17 | 6 | 100 | -0.94 |
| Class 5 | 56 | 16 | 50 | 100 | -0.44 |
| Class 6 | 9 | 25 | 7 | 77.77 | -0.91 |
| Class 7 | 45 | 20 | 45 | 100 | -0.55 |
| Class 8 | 54 | 17 | 48 | 96 | -0.16 |
| Class 9 | 10 | 26 | 6 | 60 | -0.90 |
| Class 10 | 373 | 22 | 31 | 62 | 2.73 |

The optimization results in a mean time of 19.5 seconds giving a mean accuracy are 81.69% with mean Redundancy Factor -0.124.

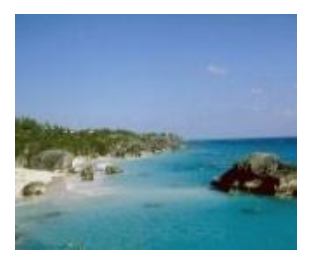
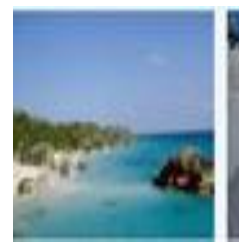
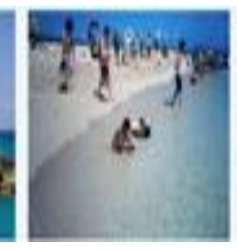
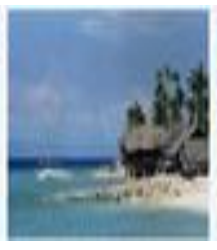
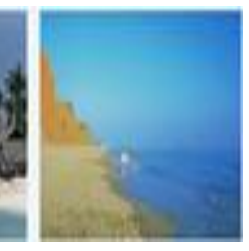

**Fig 9: Illustration of query and retrieval using Optimized Approach method**

The comparison between the individual approach, combined approach and the optimized approach using the parameters can be showed as below:

**Table 9. Comparison between the individual approach, combined approach and the optimized approach**

| Parameters | Individual Approach | Combined Approach | Optimized Approach |
|---|---|---|---|
| **Mean Time (sec)** | 12.01 | 51.3 | 19.5 |
| **Mean Accuracy (%)** | 42.05 | 91.51 | 81.69 |
| **Mean RF** | 2.65 | 0.90 | -0.124 |

So, we can clearly see that the optimized approach gives a much more balanced performance in terms all three parameters. While keeping the mean time taken for image retrieval below 20 seconds, we can achieve an accuracy of 81.69% with a RF of -0.124.

# 9. EFFECT BY CROPPING

Cropping of Images can be done when specific information is to be queried from within an image. For example, when we have a picture which has a flower in the wild, the image will be filled with green back ground and the color of the flower and we want to the results to be the similar flowers. Since the image has lot of information, this will confuse the system and it would yield irrelevant results. It is better to crop out the flower as a specific region of interest and then search the database in this situation. This gives the scope for 'query modification' to the user.

## 9.1 Images used for Sample Querying

Three images were used; Image 1 consists of a child with background of plants and trees, the desired result for this is an image of any human. The results were given by Color Moments and Local Color Histogram.

Image 2 consists of a flower with green background of plants. The desired result is an image which consists of a flower. The cropped images consist of the flower alone with little background information on the edges of the image. The results are given by using Color Moments, Global Color Histogram and Geometric Moment.

Image 3 consists of a horse and a man with prevalent brown background, the desired result for this is an image of horse. The results were given by Color Moments and Local Color Histogram.

**Before Cropping** **After Cropping**

**Image 1**

**Image 2**

**Image 3**

**Fig 10: Before cropping and after cropping for 3 images**

## 9.2 Results without Cropping

**Table 10.Results obtained without cropping**

| Image | Images retrieved | Time (sec) | Relevant Images | Accuracy (%) |
|---|---|---|---|---|
| 1 | 60 | 20 | 13 | 26 |
| 2 | 124 | 35 | 0 | 0 |
| 3 | 20 | 15 | 3 | 15 |

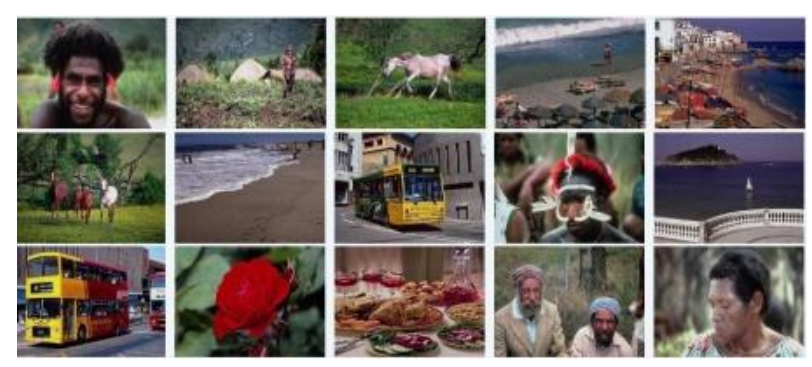

**Fig 11: Result for Image 1 before Cropping**

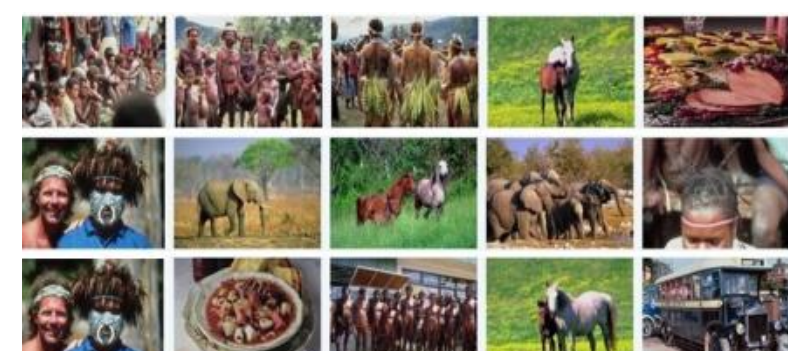

**Fig 12: Result for Image 2 before Cropping**

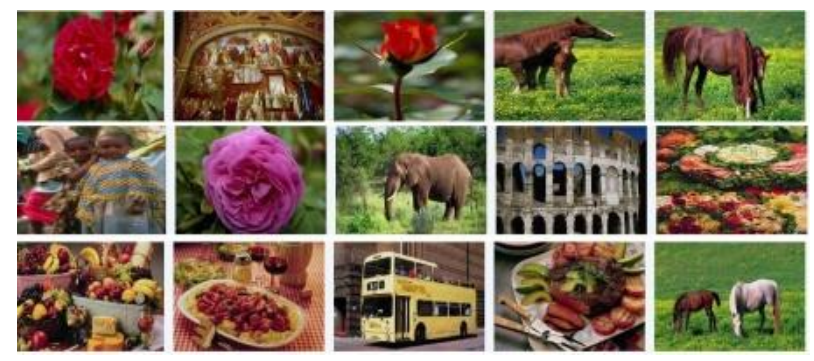

**Fig 13: Result for Image 3 before Cropping**

## 9.3 Results with Cropping

**Table 11.Results obtained with cropping**

| Image | Images retrieved | Time (sec) | Relevant Images | Accuracy ( % ) |
|---|---|---|---|---|
| 1 | 280 | 30 | 29 | 58 |
| 2 | 7 | 23 | 2 | 28 |
| 3 | 91 | 20 | 19 | 38 |

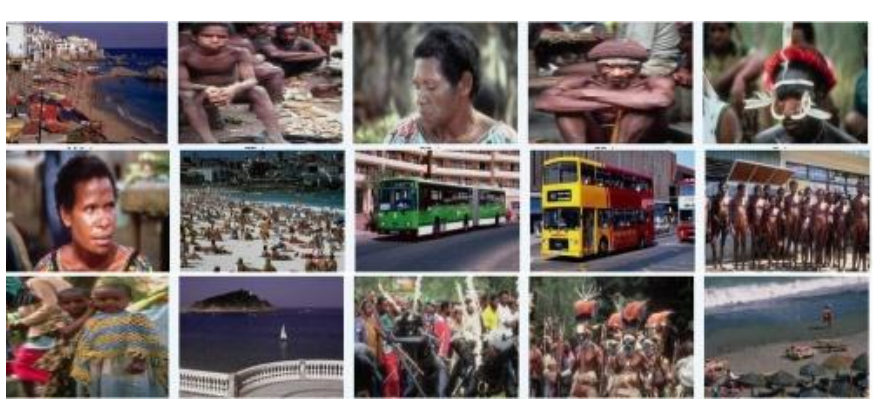

**Fig 14: Result for Image 1 after Cropping**

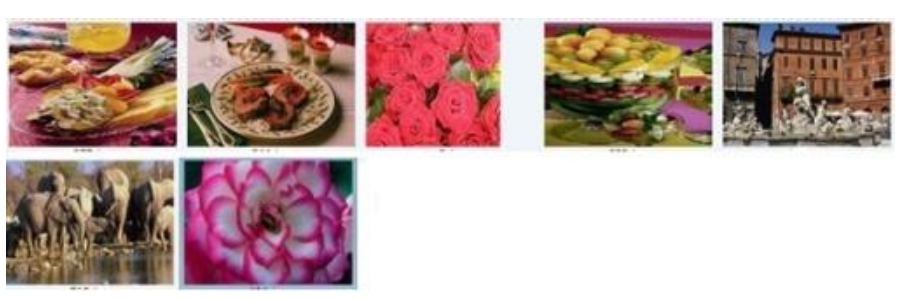

**Fig 15: Result for Image 2 after Cropping**

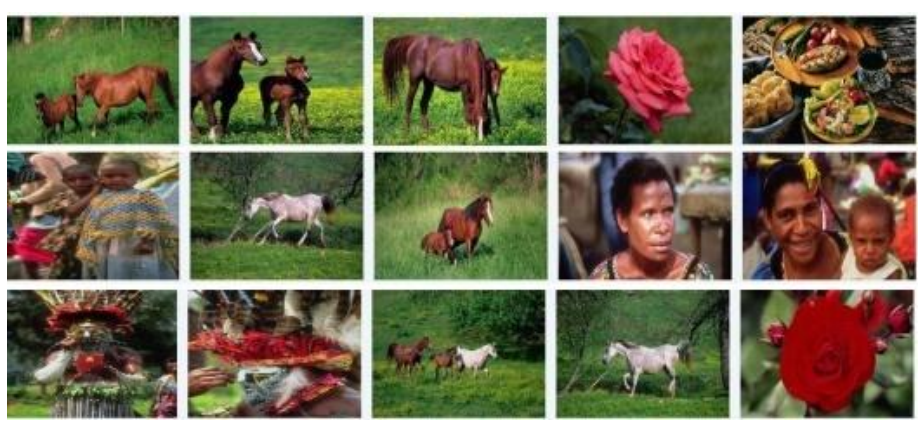

**Fig 16: Result for Image 3 after Cropping**

## 9.4 Improvement Using Cropping

We can see the improvement in accuracy by 28% on an average, from the two tables above. Cropping the image reduces the unwanted information of an image and thus helps increasing accuracy for the desired result.

## 10. CONCLUSION AND FUTURE SCOPE

We have successfully shown the comparative analysis of the various feature extraction techniques and their drawbacks when used individually. We have proposed a solution by optimizing the techniques for each class of images, resulting in an 'adaptive' retrieval system which results in a balanced performance in terms of image retrieval time, accuracy and redundancy factor. Such a system can effectively recognize the class of the image query given by the user and can produce the best results according to it. To enhance the adaptability of the system, we have also proposed the image cropping feature to identify the user's region of interest in a specific image and thus, resulting in more precise and personalized search results.

This system can be integrated with the powerful Relevance Feedback technique [12]-[15] to improve the performance over a period of time. The adaptability can be enhanced by reducing the number of iterations by using the navigation patterns of the user queries [16]. The image cropping feature can be integrated with Google to augment the results by providing the user with multiple sources of information relevant to the query image. For example, if the cropped query image is a picture of a shopping mall, integrating it with Google Maps can provide maps and routes to the place of interest along with other relevant images and information about it. The retrieval performance can be further improved by using a 'Text and Image' query system as compared to a text-only or image-only query system, which can take advantage of the keyword annotations. The results from the keyword annotations and image retrieval can then be matched using the feature extraction techniques to present an optimized set of results to the user.

## 11. ACKNOWLEDGMENTS

Our sincere thanks to Shikhar, Shraddha and Vijay for providing us information regarding the CBIR systems in-place today.

## AUTHOR BIOGRAPHIES

**AmanChadha**(M'2008) was born in Mumbai (M.H.) in India on November 22, 1990. He is currently pursuing his graduate studies in Electrical and Computer Engineering at the University of Wisconsin-Madison, USA. He completed his B.E. in Electronics and Telecommunication Engineering from the University of Mumbai in 2012. His special fields of interest include Signal and Image Processing, Computer Vision (particularly, Pattern Recognition) and Processor Microarchitecture. He has 10 papers in International Conferences and Journals to his credit. He is a member of the IETE, IACSIT and ISTE.

**SushmitMallik** (M'2008) was born in Kolkata (W.B.) in India on October12, 1990. He is currently pursuing his graduate studies in Electrical Engineering at North Carolina State University, Raleigh, USA He completed his B.Tech. in Electronics and Communication Engineering from SRM University in 2012.Previously, he was a visiting student at the University of Wisconsin-Madison, USA in 2011 and a Student Research Assistant at the University Of Hong Kong in 2012. His fields of interest include Nanoelectronics, Optoelectronic devices and Robotics.

**RavdeepJohar** (M'2008) was born in Bokaro (J.H.) in India on July 16, 1991. He completed his B.Tech.in Computer Science Engineering from SRM University in 2012. Previously, he was a visiting student at the University of Wisconsin-Milwaukee, USA in 2011.His fields of interest include Computer Graphics, Computer Vision and Programming Languages.